%% file: samplepaper.tex
\documentclass[runningheads]{llncs}
\usepackage[T1]{fontenc}

\usepackage{graphicx}
\usepackage{tabularx}
\usepackage{hhline}
\usepackage{booktabs}
\usepackage{hyperref}
\usepackage{amsmath}
\usepackage{amssymb}
\usepackage{bbm}
\usepackage{fancyhdr,graphicx,amsmath,amssymb}
\usepackage[ruled,vlined]{algorithm2e}
\usepackage{subfiles}

\begin{document}
\title{Subgroup-Specific Risk-Controlled Dose Estimation in Radiotherapy}

\author{Paul Fischer$^*$\inst{1,2} \and Hannah Willms$^*$\inst{1} \and Moritz Schneider\inst{3} \and Daniela Thorwarth\inst{1,3} \and Michael Muehlebach\inst{4} \and Christian F. Baumgartner\inst{1,2}}

\authorrunning{P. Fischer et al.}

\institute{Cluster of Excellence -- ML for Science, University of Tübingen, Germany\\ 
\and
Faculty of Health Sciences and Medicine, University of Lucerne, Switzerland\\
\and
Section for Biomedical Physics, Department of Radiation Oncology, University of Tübingen, Germany\\
\and
Max Planck Institute for Intelligent Systems, Tübingen, Germany\\
}

\maketitle
\def\thefootnote{*}\footnotetext{Contributed equally}\def\thefootnote{\arabic{footnote}}
\begin{abstract}
\setcounter{footnote}{0}
Cancer remains a leading cause of death, highlighting the importance of effective radiotherapy (RT). Magnetic resonance-guided linear accelerators (MR-Linacs) enable imaging during RT, allowing for inter-fraction, and perhaps even intra-fraction, adjustments of treatment plans. However, achieving this requires fast and accurate dose calculations. While Monte Carlo simulations offer accuracy, they are computationally intensive. Deep learning frameworks show promise, yet lack uncertainty quantification crucial for high-risk applications like RT. Risk-controlling prediction sets (RCPS) offer model-agnostic uncertainty quantification with mathematical guarantees. However, we show that naive application of RCPS may lead to only certain subgroups such as the image background being risk-controlled. In this work, we extend RCPS to provide prediction intervals with coverage guarantees for multiple subgroups with unknown subgroup membership at test time. We evaluate our algorithm on real clinical planing volumes from five different anatomical regions and show that our novel subgroup RCPS (SG-RCPS) algorithm leads to prediction intervals that jointly control the risk for multiple subgroups. In particular, our method controls the risk of the crucial voxels along the radiation beam significantly better than conventional RCPS. 


\end{abstract}
\section{Introduction}
Cancer remains one of the leading causes of death for people under the age of 70.
Radiotherapy (RT) has been proven to be a critical treatment modality for various tumour entities.
Recent advancements in medical imaging have led to the development of the magnetic resonance-guided linear accelerator (MR-Linac), which integrates MR imaging with RT~\cite{hall2019MRlinac}. Aligning the computed tomography (CT) planing volume to an MR scan acquired at the beginning of each treatment fraction, and re-evaluating the treatment plan based on the transformed CT volume allows to better adjust the plan to the patient's inter-fraction changes.
This holds the potential for more precise tumor targeting, enhancing treatment outcomes.
However, in order to successfully realize these benefits, fast and accurate dose deposition calculations are needed. Current algorithms are based on Monte Carlo simulations, which provide highly accurate results (\cite{bol2012fast}).
However, these calculations are computationally intensive, taking minutes to hours.
Therefore recent research has focused on optimizing treatment planning efficiency~\cite{randall2022towards}.
Deep learning (DL) frameworks have shown encouraging results in dose estimation for RT, with fast calculation time and accurate predictions~\cite{kontaxis2020deepdose,martinot2021high,neishabouri2021long,tsekas2021deepdose}. However, despite RT's high-risk nature, prior approaches have not adequately addressed risk assessment within DL-based dose estimation.

A natural approach for assessing the risk associated with a prediction is quantifying the prediction uncertainty. Recent years have seen the development of various methods for uncertainty quantification in medical imaging. Examples include deep ensembles~\cite{lakshminarayanan2017simple}, Monte Carlo dropout~\cite{kendall2015bayesian}, or approaches based on variational autoencoders~\cite{baumgartner2019phiseg,fischer2023uncertainty,kohl2018probabilistic}. A major limitation of those techniques is that they do not provide any guarantees about the usefulness or correctness of the uncertainty estimates. Recently, risk-controlling prediction sets~\cite{bates2021distribution} (RCPS) has gained popularity as a simple, model-agnostic strategy to adapt heuristic notions of uncertainty into uncertainty measures with guarantees. RCPS allows to construct a set of predictions with a guarantee that the correct solution is inside this set with a user-defined probability. 
Such prediction sets can indicate poor model performance through excessively large intervals, revealing that the models may not be acceptable for certain high-risk applications.

While RCPS has already seen successful adoption in medical image analysis (e.g. \cite{angelopoulos2022image}), a remaining limitation is that it can only provide guarantees on a global level. There are many situations where we are interested in obtaining guarantees for different subgroups of our population, or different image regions. For instance, in RT, we would like the method to be calibrated along the beam as well as the background. If there is an imbalance between different subgroups (e.g. more background voxels) naive application of RCPS will focus mostly on the majority group (e.g. background) and fail to meet the guarantees for the minority subgroup (e.g. the beam). Calculating RCPS for the different subgroups separately is only a solution if the subgroup is known at test time. However, if this is not the case, as in our RT example, it is not possible to determine the correct RCPS model to use for prediction. 

\begin{figure}[t]
    \centering
    \includegraphics[width=\textwidth]{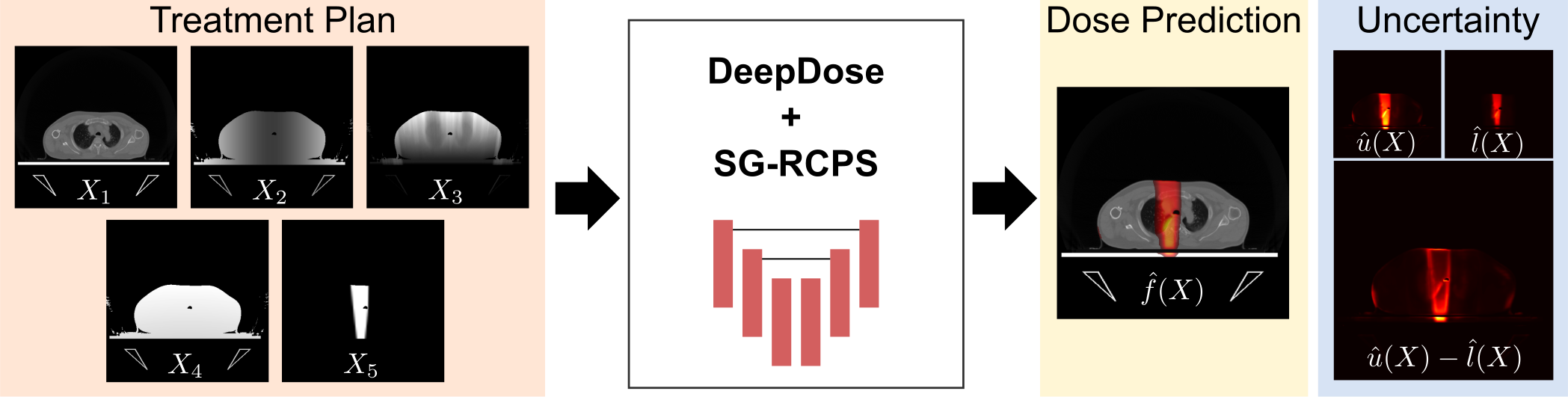}
    \caption{\textbf{Overview.} We use the DeepDose network~\cite{kontaxis2020deepdose} to convert a personalized RT plan defined by input CT scan, beam center distance, radiological depth, source distance map, and beam shape ($X_1$ to $X_5$) to a voxel-wise dose prediction $\hat{f}(x)$. Extending DeepDose by our novel subgroup risk-controlled prediction sets algorithm (SG-RCPS) allows to obtain a calibrated upper and lower bound for the dose ($\hat{u}(X)$ \& $\hat{l}(X)$), as well as the voxel-wise size of the interval ($\hat{u}(X)-\hat{l}(X)$) which serves as final uncertainty measure.}
    \label{fig:method}
\end{figure}

In this paper, we address this problem by proposing a novel calibration algorithm for RCPS that takes into account subgroups and can provide subgroup as well as global guarantees. Our contributions are:
\begin{enumerate}
    \item The first application of uncertainty quantification in neural network-based dose estimation for RT.
    \item A novel algorithm that yields mathematical guarantees for uncertainty intervals for subgroups in the dataset.
    \item A quantitative and qualitative evaluation of the algorithm on RT dose prediction on a real-world multi-organ dataset.
\end{enumerate}

\section{Methods}

The RCPS framework~\cite{angelopoulos2022image,bates2021distribution} ensures that a set-valued predictor $\mathcal{T}$ maintains a \emph{risk} below a user-specified level $\alpha$ with a user-defined probability of $1 - \delta$. In regression problems such as ours, the prediction set is often a prediction interval characterised by a lower and an upper bound value. The risk $\mathcal{R}(\mathcal{T}) = \mathbb{E}[L(Y, \mathcal{T}(X))]$ is defined through a loss function $L$ tailored to the application, which encodes a notion of consequence if the desired property is not fulfilled. In this work, we demonstrate the application of RCPS to DL-based dose prediction and extend the method to ensure risk guarantees for multiple subgroups.

In the following, we will first describe our dose estimation framework (Sec.~\ref{sec:deepdose}). We will then show how heuristic prediction intervals can be obtained using quantile regression (Sec.~\ref{sec:quantile-regression}). Next, we will discuss how to define the concept of risk, and how the heuristic prediction intervals can be adjusted to control the risk with the desired levels (Sec.~\ref{sec:rcps}). Lastly, we will describe our novel subgroup RCPS algorithm which allows controlling the risk for multiple subgroups without knowledge of the subgroup membership at test time (Sec.~\ref{sec:subgroup-calibration}). 

\subsection{Dose Estimation using DeepDose}
\label{sec:deepdose}
In order to construct a voxel-wise dose predictor $\hat{f}$ we build on the previously proposed DeepDose network~\cite{kontaxis2020deepdose,tsekas2021deepdose}, which is in turn derived from a 3D UNet architecture~\cite{cciccek20163d}. DeepDose takes the beam shape, center beam line distance, source distance, CT image and radiological depth as input $X \in \mathbb{R}^{5 \times W \times H \times D}$ and outputs a predicted dose $\hat{f}(X)_i \in \mathbb{R}$ for each voxel $i$ (see~Fig.\ref{fig:method}). The network is trained  with a highly accurate Monte Carlo simulation as ground truth $Y_i$ for each voxel. For notational clarity, we will omit the index $i$ in the following.

\subsection{Heuristic Dose Prediction Intervals using Quantile Regression}
\label{sec:quantile-regression}
To extend the DeepDose network by the ability to provide prediction intervals, we adopt a voxel-wise quantile regression approach~\cite{angelopoulos2022image}. 
Specifically, we add two additional output channels $\tilde{l}(X)$ and $\tilde{u}(X)$ to estimate the voxel-wise upper and lower bound, respectively. Similar to \cite{angelopoulos2022image}, we train the two additional network heads using pinball losses, which allow to estimate a specific quantile. 
For a general feature $x$, label $y$, and quantile $\beta$, the pinball loss is given by
\begin{equation}
\begin{split}
    \mathcal{L}_\beta(\hat{q}_\beta(x), y) &= (y-\hat{q}_\beta(x))\beta\mathbbm{1}_{\{y>\hat{q}_\beta(x)\}} + \\
    &\hspace{2cm}(\hat{q}_\beta(x)-y)(1-\beta)\mathbbm{1}_{\{y\leq\hat{q}_\beta(x)\}},
\end{split}
\end{equation}
where $\hat{q}_\beta(x)$ is the corresponding quantile estimator and $\mathbbm{1}$ denotes the indicator function. We use $\hat{f}(X) - \tilde{l}(X)$ and $\hat{f}(X) + \tilde{u}(X)$ for the respective lower and upper quantile estimators. 
Our overall training objective $\mathcal{L}$ is comprised of losses for the upper and lower quantiles as well as the standard MSE loss for the point prediction 
\begin{equation}
\label{eq:heuristic-prediction-set}
    \mathcal{L} = \mathcal{L}_{\alpha/2}(\tilde{l}(X), Y) + \mathcal{L}_{1-\alpha/2}(\tilde{u}(X), Y + \text{MSE}(Y, \hat{f}(X)),
\end{equation}
where each loss is only applied to the corresponding head.
This gives our framework the ability to not only output a per-voxel dose prediction $\hat{f}(X)$, but also a heuristic prediction interval 
\begin{equation}
\mathcal{T}(X) = [\hat{f}(X) - \tilde{l}(X), \hat{f}(X) + \tilde{u}(X)].
\end{equation}

\subsection{RCPS for Radiotherapy Dose Estimation}
\label{sec:rcps}
 
We now show how the RCPS framework~\cite{angelopoulos2022image,bates2021distribution} can be used to obtain dose prediction intervals that are guaranteed to keep a \emph{risk} below a user-specified level.
First, we define the risk of $\mathcal{T}$ as the predicted interval \emph{not} containing the ground truth dose $Y$
\begin{equation}
    \mathcal{R} (\mathcal{T}) = \mathbb{E} \left[ \mathbbm{1}_{\{Y\not\in \mathcal{T}(X)\}} \right]=\text{Pr}(Y\not \in \mathcal{T}(X)).
\end{equation} 
We then define new lower and upper bounds by scaling them with a non-negative factor $\hat{\lambda}$
\begin{equation}
\label{eq:corrected-bounds}
\hat{l}(X)=\hat{\lambda}\tilde{l}(X) \quad \text{and} \quad \hat{u}(X)=\hat{\lambda}\tilde{u}(X).
\end{equation}
RCPS provides a strategy to choose $\hat{\lambda}$ based on a calibration dataset such that $\mathcal{R}(\mathcal{T}) \leq \alpha$ with a probability of at least $1-\delta$ on future test data under the assumption of exchangeability of the test and calibration sets. We use $\alpha = \delta = 0.1$ for all experiments, meaning that with a probability of at least 90\%, a minimum of 90\% of the ground truth dose depositions should be contained in the predicted dose interval.

Since the calibration data is only a random sample of the data distribution, $\hat{\lambda}$ cannot be chosen by simply minimising the risk on the calibration set. Rather a point-wise \emph{upper confidence bound (UCB)} $\mathcal{R}^+ : \text{Pr}(\mathcal{R}(\lambda)) \leq \mathcal{R}^+$ must be obtained. This UCB accounts for the calibration sample size and the desired probability of the guarantee holding on future data. Following \cite{angelopoulos2022image} we use the Hoeffding bound~\cite{hoeffding1994probability} to define $\mathcal{R}^+$ as
\begin{equation}
\label{eq:ucb}
R^+(\lambda) = \frac{1}{nWHD} \sum_{k=1}^n \#\{\mathcal{T}_\lambda(X_k)\not\in Y_k\} + \sqrt{\frac{1}{2n} \log \frac{1}{\delta}} ~ \text{.}
\end{equation}

In the original RCPS approach, Bates et al.~\cite{bates2021distribution} proposed a greedy optimization algorithm for obtaining the smallest possible prediction interval still fulfilling the desired guarantees. It requires initializing $\hat{\lambda}$ with a very large value and reducing it until $\mathcal{R}^+$ falls under the desired risk level, that is,
\begin{equation}
\label{eq:optimal-lambda}
    \hat{\lambda} = \min \left\{ \lambda : \hat{R}^+(\lambda) \leq \alpha \right\}.
\end{equation}
This procedure produces a dose interval predictor
\begin{equation}
\mathcal{T}_{\hat{\lambda}}(X) = [\hat{f}(X) - \hat{\lambda}\tilde{l}(X), \hat{f}(X) + \hat{\lambda}\tilde{u}(X)],
\end{equation}
for the predicted dose that satisfies the desired risk-properties \textit{on average} for all voxels of the test data distribution. However, it does not offer conditional guarantees for individual data subgroups. For instance, as we will show in Sec.~\ref{sec:experiments} a naive application of RCPS results in the voxels along the radiation beam violating the desired guarantees. 

\subsection{Risk-Controlled Prediction Sets with Multiple Subgroups} 
\label{sec:subgroup-calibration}
If our dataset comprises $M$ imbalanced subgroups, using Eq.~\ref{eq:optimal-lambda} to obtain $\hat{\lambda}$ provides guarantees for the overall dataset but not for each individual subgroup. This can lead to a systematic miscalibration of uncertainty intervals for under-represented subgroups. If the subgroups are known at test time, this problem can be addressed by calibrating for each subgroup separately, by using a subgroup-specific parameter $\hat{\lambda}_z$ during test time. In our RT application, we are interested in calibrated uncertainty quantification in the area of the beam as well as the background, which also receives small dose intensities. While we know the absorbed dose during training, we do not have direct access to this information during testing. Hence, naive application of RCPS yields good calibration overall, but not in the critical area of the beam. 

Therefore, we propose an extension of the risk-controlling framework for scenarios where the subgroup membership $Z$ is unknown at test time. Our proposed extension provides the same guarantees as RCPS for each individual subgroup. That is for every subgroup $Z$ in our dataset, it holds that, at a risk level of $\alpha$, the ground truth is included with a probability of at least $1-\delta$.

In order to achieve the desired guarantees we reformulate the risk conditioned on the subgroups $Z$ as follows
\begin{equation}
\mathcal{R} (\mathcal{T}) =\text{Pr}(Y\not\in\mathcal{T}(X)) = \mathbb{E}_Z [ \mathbb{E}_{XY|Z} \left[ \mathbbm{1}_{\{Y\not\in\mathcal{T}(X)\}} \right]] ~ \text{.}
\end{equation}
This leads to a novel subgroup risk-controlled predictions set (SG-RCPS) procedure which is summarised in Algorithm~\ref{alg1}. Similar to the original RCPS algorithm, we start off with a large $\hat{\lambda}$ that satisfies the risk for each subgroup. We then iteratively reduce the interval size until the first confidence bound of a subgroup no longer satisfies the criterion\footnote{The code is available at \url{https://github.com/paulkogni/SG-RCPS}}. A proof that Algorithm~\ref{alg1} leads to the desired subgroup guarantees is presented in Appendix A.

\begin{algorithm}[H]
\label{alg1}
\SetAlgoLined
\SetKwInOut{Input}{Input}
\SetKwInOut{Output}{Output}
\Input{Calibration sets $(X_k, Y_k)_z$, $k=1, \dots, n_z$ where $z=1, \dots, M$ indicates the subgroup; in our case we have $M=3$ for foreground, background and the combined image; risk level $\alpha$; error rate $\delta$; predictor $\hat{f}$; heuristic lower and upper interval predictions $\tilde{l}$ and $\tilde{u}$; initial max value $\lambda_\text{max}$; step size $d\lambda > 0$}
\Output{Optimal interval scaling $\hat{\lambda}$}
 $\lambda \leftarrow \lambda_\text{max}$ \\
 \For{$z\gets 1$ \KwTo $M$}{
  $UCB_z \leftarrow 1$ \\
 }
 \While{$UCB_{1} \leq \alpha \ \& \dots \& \ UCB_{M} \leq \alpha$}{
  $\lambda \leftarrow \lambda - d\lambda$\\
  \For{$z\gets 1$ \KwTo $M$}{
      \For{$k\gets0$ \KwTo $n_z$}{
        $L_{k, z} \leftarrow \#\{\mathcal{T}_{\lambda} (X_{k,z})\not\in Y_{k,z}\}/WHD$ \\
      }
  $UCB_z \leftarrow \frac{1}{n_z} \sum_{k=1}^{n_z} L_{k, z} + \sqrt{\frac{1}{2n_z} \log \frac{1}{\delta} }  $ \\
  }
 }
 $\hat{\lambda} \leftarrow \lambda + d\lambda$
 \caption{Pseudocode for SG-RSPC}
\end{algorithm}

\section{Experiments and Results}
\label{sec:experiments}
\subsection{Dataset}
\label{subsec:data}
To assess the performance of our model, we trained and tested it on a dataset containing CT data and RT treatment plans of 125 patients obtained from patients at the Department of Radiation Oncology at the University of Tübingen. The study was approved by the institutional review board and all patients gave written informed consent (NCT04172753).

The training dataset comprises four anatomical entities: prostate, liver, breast (mamma), and head and neck (HN). The test and calibration datasets contained data from the same four tumour entities as the training dataset, along with lymph nodes, as an additional out-of-domain (OOD) entity.
Testing the neural network on an OOD entity allowed us to assess whether the calibration is able to generalize, which is highly desirable in real-world scenarios.
For each patient, we extracted multi-leaf collimator (MLC) segments, resulting in a total of 6638 segments. For validation, we randomly selected 20 diverse prostate segments from the dataset.
For calibration, we randomly selected three random segments from one patient for each entity. More detailed information about the data splits is provided in Appendix B. To train the DeepDose network, each segment was divided into patches of size $5\times 32 \times 32 \times 32$.  The ground truth dose estimations were generated using Monte Carlo simulations with the EGSnrc open-source software package \cite{friedel2019development,egsnrc}.  For inference, we performed predictions at the patch level and combined them in a sliding window fashion.

\subsection{Findings}
\begin{figure}[t]
    \centering
    \includegraphics[width=\textwidth]{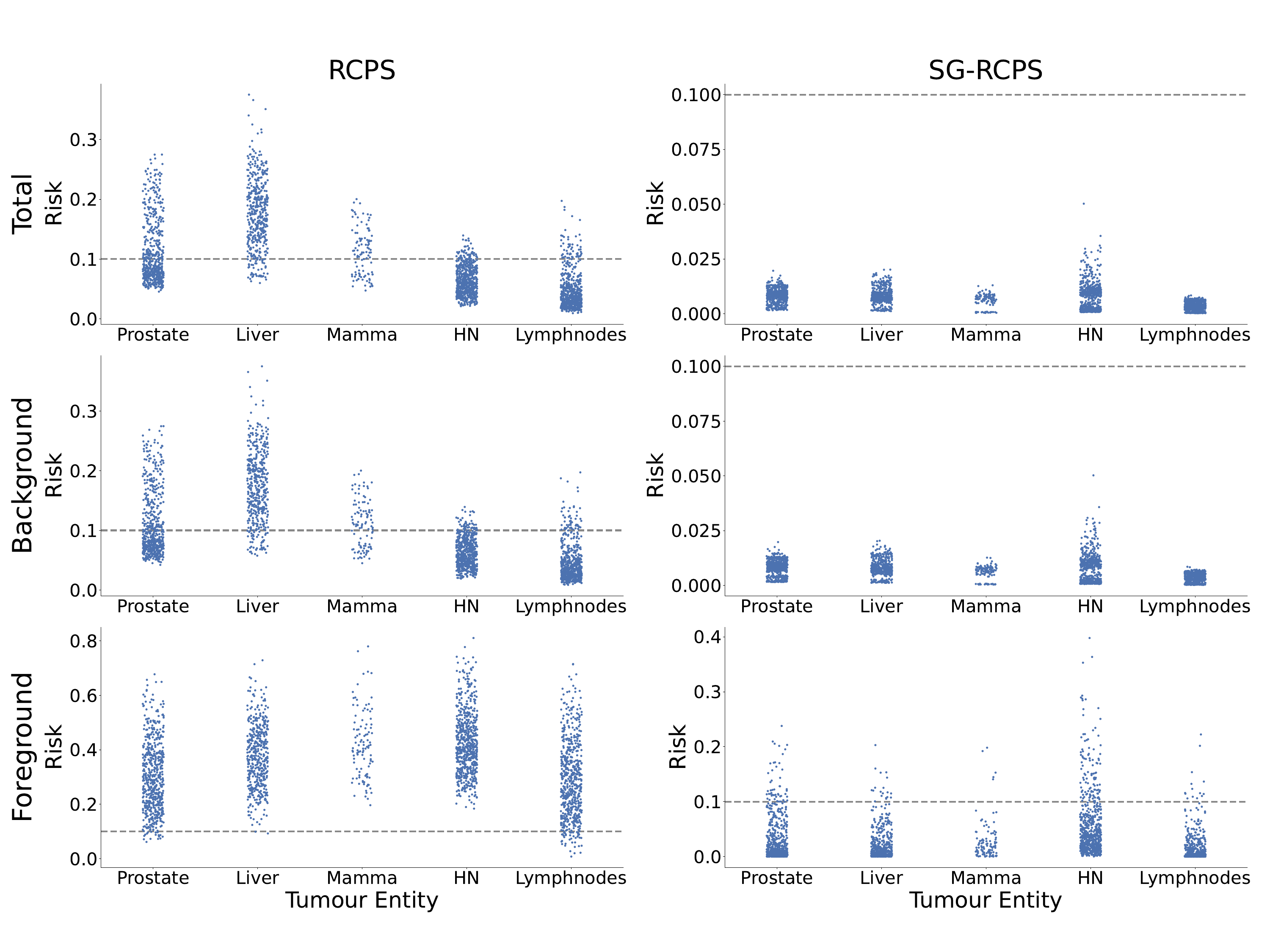}
    \caption{Tumor-specific risks for the original calibration method (left) and our method (right) for the total image (top row), the background radiation (middle row) and foreground radiation (bottom row).} 
    \label{fig:risks}
\end{figure}

\begin{table}[t]
\caption{\textbf{Quantitative results.} Empirical risks for RCPS and SG-RCPS averaged over all segments. Controlled risks with $\alpha\leq 0.1$ for more than $1-\delta$ of the cases are highlighted in bold.}
\begin{center}
\vspace{-2mm}
\resizebox{\textwidth}{!}{%
\begin{tabular}{lcccccccccccc}
& \multicolumn{2}{c}{Prostate} & \multicolumn{2}{c}{Liver} & \multicolumn{2}{c}{Mamma} & \multicolumn{2}{c}{HN} & \multicolumn{2}{c}{Lymph.} & &\\
\cmidrule(lr){2-3} \cmidrule(lr){4-5} \cmidrule(lr){6-7} \cmidrule(lr){8-9} \cmidrule(lr){10-11}
Method & \multicolumn{1}{c}{RCPS} & \multicolumn{1}{c}{SG-RCPS} & \multicolumn{1}{c}{RCPS} & \multicolumn{1}{c}{SG-RCPS} & \multicolumn{1}{c}{RCPS} & \multicolumn{1}{c}{SG-RCPS} & \multicolumn{1}{c}{RCPS} & \multicolumn{1}{c}{SG-RCPS} &\multicolumn{1}{c}{RCPS} & \multicolumn{1}{c}{SG-RCPS}  \\
\toprule
Total image & 0.415 & \textbf{0.0} & 0.898 & \textbf{0.0} & 0.620 & \textbf{0.0} & \textbf{0.089} & \textbf{0.0} & \textbf{0.094} & \textbf{0.0}\\
Backgr. rad. & 0.403 & \textbf{0.0} & 0.896 & \textbf{0.0} & 0.610 & \textbf{0.0} & \textbf{0.083} & \textbf{0.0} & \textbf{0.094} & \textbf{0.0}\\
Foregr. rad. & 0.970 & \textbf{0.084} & 0.996 & \textbf{0.041} & 1.0 & \textbf{0.048} & 1.0 & 0.176 & 0.934 & \textbf{0.024}\\
\bottomrule
\end{tabular}
}
\end{center}
\label{tab:results}
\vspace{-5mm}
\end{table}

We trained a DeepDose network using training data from all four entities. To evaluate the model's dose estimation performance we calculated the 3mm/3\% gamma pass rate ($\gamma$-PR)  criterion and observed a $\gamma$-PR of 98.9\%. 

We then compared the two calibration strategies discussed above: RCPS, and our proposed subgroup RCPS (SG-RCPS). We considered three subgroups: the beam foreground and background determined by thresholding the ground truth dose (see Fig.~\ref{fig:method}), and the whole image. We used a target risk level of $\alpha = 0.1$, and an error rate of $\delta = 0.1$. 

Fig.~\ref{fig:risks} and Tab.~\ref{tab:results} show the empirical risk for all segments in the test set grouped by entity. The empirical risk for each segment is defined as proportion of ground truth doses not contained in the predicted interval. Based on our risk settings we expect at most 10\% of the segments to fall above the specified risk of 10\%.

From Fig.~\ref{fig:risks} it can be seen that the calibration is dominated by the background class. We found that the normal RCPS algorithm only controlled for the risk in the head \& neck, and lymph node entities when considering the total image, and the background only. However, the risk was not controlled to the desired levels in the foreground subgroup (i.e. the beam) for any of the entities. Interestingly, the risks for liver, prostate as well as mamma are not controlled even in the total image. This is likely caused by a mismatch in the proportion of background voxels in the calibration and the test set. 

Our proposed SG-RCPS algorithm was able control the risk substantially better for all anatomical areas. There were no dose predictions outside the predicted interval when considering the total image and the background only. We note that because our algorithm estimates a single $\hat{\lambda}$ that controls the risk for all subgroups jointly, the estimates for the total image and background group were more conservative.
When considering the foreground (i.e. the beam) only, we found that all entities except head \& neck were risk-controlled with the desired levels. As can be seen in Tab.~\ref{tab:results} the empirical risk for the head \& neck entity fell slightly short of the desired levels. 
Notably, the risk for out-of-distribution entity, lymph nodes, was also well controlled. Predicting risk-controlled intervals is particularly important for the foreground, as the the beam is where the most accurate uncertainty estimation is required. 

\begin{figure}[t]
    \centering
    \includegraphics[width=\textwidth]{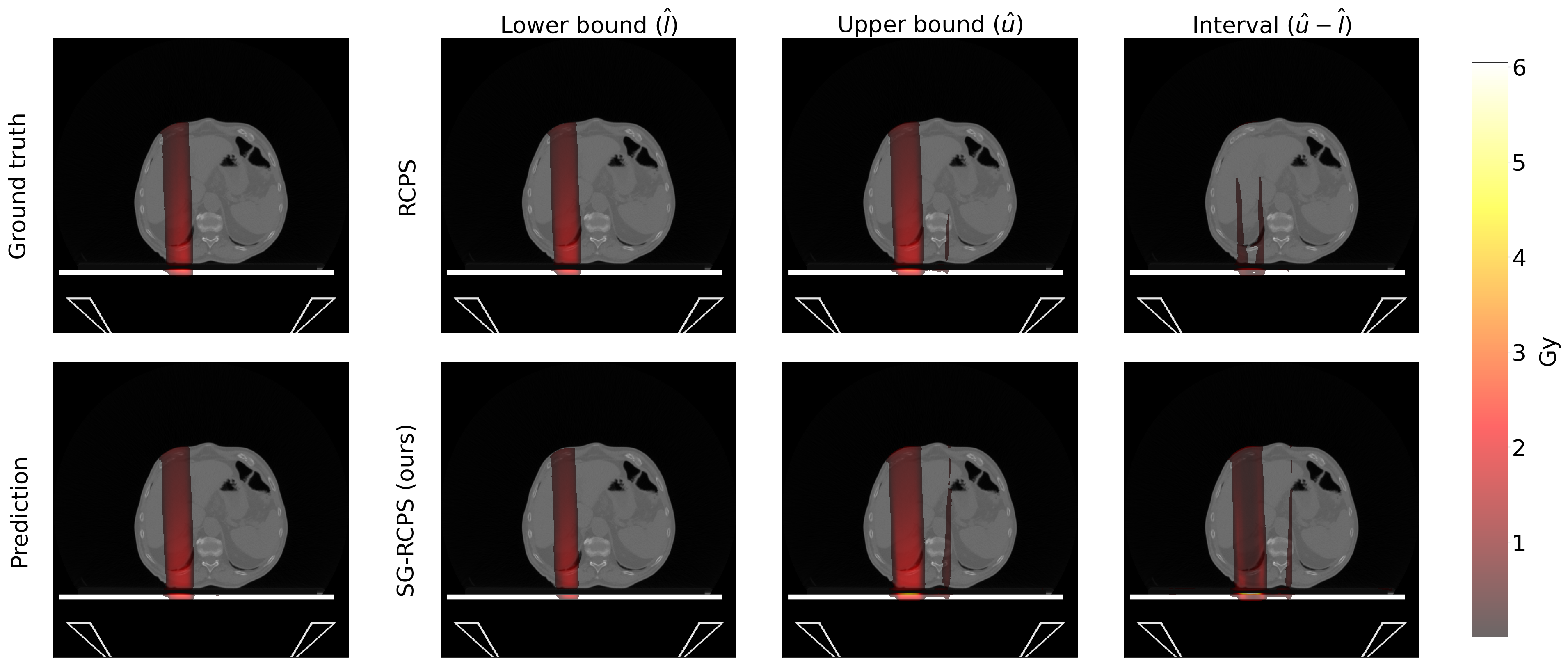}
    \caption{A representative example for a liver tumor visualizing the qualitative differences between the uncertainty intervals for the non-subgroup-specific calibration and our method. The uncertainty intervals provided by our method are significantly wider ($p < 0.001$) than the ones generated by classical RCPS. All values are given in Gray (Gy).}
    \label{fig:quali}
\end{figure}

A qualitative example of dose predictions and prediction intervals for RCPS and SG-RCPS is shown in Fig.~\ref{fig:quali}.

\section{Conclusion}
We have proposed subgroup RCPS, an extension of the RCPS algorithm allowing to control risk for multiple subgroups with unknown subgroup membership at test time. We validated our method on a clinical RT dataset comprising five anatomical entities. Our results demonstrate that in case of imbalances between subgroups our method substantially improves calibration for individual subsets. Specifically, in contrast to regular RCPS, our SG-RCPS approach allows to control the risk for the beam \textit{and} the background thereby increasing safety and trustworthiness in this high-risk application. A potential drawback of our method is that it requires a separate calibration set for each subgroup. Additionally, this method usually yields more conservative prediction intervals. 
In the future, we will apply this algorithm to datasets that include other under-represented subgroups, such as ethnicity or gender.

\begin{credits}
\subsubsection{\ackname}  This work was supported by the Excellence Cluster 2064 ``Machine Learning --- New Perspectives for Science'', project number 390727645 and received funding from the German Research Council under DFG Grant No. ZI 736/2-1. The authors thank the International Max Planck Research School for Intelligent Systems (IMPRS-IS) for supporting Paul Fischer. The authors acknowledge support by the state of Baden-Württemberg through bwHPC (INST39/963-1 FUGG bwForCluster NEMO) and through the Research and Training Network "AI4MedBW". The authors thank Julius Vetter for his valuable feedback on this work.

\subsubsection{\discintname}
The authors have no competing interests to declare. 
\end{credits}


\bibliographystyle{splncs04}
\bibliography{bibliography}

\newpage

\subfile{supplemental.tex}
\end{document}

%% file: supplemental.tex
\begin{center}
{\Large \bf Supplemental Materials \\ \vspace{2mm}
}
\end{center}

\setcounter{figure}{0} 

\section{Derivation of the SG-RCPS algorithm}
\paragraph{Without subgroups:} We start by summarizing the situation without subgroups. Let $X$ be the set of features and $Y$ the corresponding set of responses. Additionally, let $\mathcal{T}$ be a predictor with $\mathcal{T}: X \rightarrow \hat{Y}$ where $\hat{Y}$ is the space of sets that include different responses $Y$. The risk of $\mathcal{T}$ is defined as 
\begin{equation*}
    \mathcal{R} (\mathcal{T}) = \mathbb{E} \left[ \mathbbm{1}_{\{Y\not\in \mathcal{T}(X)\}} \right]=\text{Pr}(Y\not \in \mathcal{T}(X)),
\end{equation*} 
where the expectation is taken over the distribution of $X$ and $Y$ on the calibration data set. Using the RCPS framework we can construct a predictor $\mathcal{T}$ such that $\mathcal{R}(\mathcal{T}) \leq \alpha$ with a probability of at least $1-\delta$. We note that the indicator function is bounded, which means that the risk is guaranteed to be bounded by 
\begin{equation}
\begin{split}
\mathcal{R}(\mathcal{T}) &= 
\mathbb{E}[ \mathbbm{1}_{\{Y\not\in\mathcal{T}(X)\}}]\\
               &= \mathbb{E}[\mathbbm{1}_{\{Y\not\in\mathcal{T}(X)\}} | \mathcal{R}(\mathcal{T}) \leq \alpha ] \cdot \text{Pr}(\mathcal{R} (\mathcal{T}) \leq \alpha) \\
               & \hspace{4mm}  +     \mathbb{E}[\mathbbm{1}_{\{Y\not\in\mathcal{T}(X)\}} | \mathcal{R}(T) > \alpha ] \cdot (1 - \text{Pr}(\mathcal{R} (\mathcal{T}) \leq \alpha)) \\
               & \leq \alpha + \delta.\label{eq:single}
\end{split}
\end{equation}
\paragraph{With subgroups:} We model the case with multiple subgroups in the dataset by introducing an additional random variable $Z$ that takes values in $\{1,\dots,K\}$ and addresses the different subgroups. For example, if $Z$ takes the value $1$, $(X,Y)$ is assumed to be distributed according to the first subgroup, if $Z$ takes the value $2$, $(X,Y)$ is distributed according to the second subgroup, etc. The value of $Z$ is unknown at test time. Algorithm~1 ensures that the risk for each subgroup is bounded via Eq.~\eqref{eq:single}, that is,
\begin{equation*}
    \mathcal{R}(\mathcal{T})=\text{Pr}(Y\not\in\mathcal{T}(X))= \mathbb{E}[ \mathbbm{1}_{\{Y\not\in\mathcal{T}(X)\}}|Z=\Bar{z}]\leq \alpha + \delta, 
\end{equation*}
for the distribution of $(X,Y)$ conditioned on the subgroup $\bar{z}$. This is due to the fact that Algorithm~1 applies the upper confidence bound arising from Hoeffding's inequality for each subgroup $\Bar{z}  \in \{1, \dots K \}$ separately. The fact that the risk of the predictor $\mathcal{T}$ is bounded by $\alpha +\delta$ (conditional on $Z$), implies that the same holds for the distribution of $(X,Y)$ during test time:
\begin{equation}
\begin{split}
    \mathcal{R} (\mathcal{T}) =\text{Pr}(Y\not\in\mathcal{T}(X))&= \mathbb{E}_{XY}\left[ \mathbbm{1}_{\{Y\not\in\mathcal{T}(X)\}} \right]  \\
                    &= \mathbb{E}_Z [ \underbrace{\mathbb{E}_{XY|Z} \left[ \mathbbm{1}_{\{Y\not\in\mathcal{T}(X)\}} \right]}_{\leq \alpha + \delta}  ] \\ 
                    & \leq \alpha + \delta.
\end{split}
\end{equation}

\newpage
\section{Summary of the dataset}

\begin{table}[h]
\centering
\caption{Number of patients and segments per tumour entity in training dataset}
\begin{tabularx}{0.5\columnwidth}{l|cccc}
\toprule
& \multicolumn{4}{c}{Training Dataset}\\
\hline
\shortstack[l]{Overall \\ Segments}  & \multicolumn{4}{c}{3958} \\
\hline
Entity       & \multicolumn{1}{c|}{Prostate} & \multicolumn{1}{c|}{Liver} & \multicolumn{1}{c|}{HN} & Mamma \\
\hline
Patients & \multicolumn{1}{c|}{40}       & \multicolumn{1}{c|}{15}    & \multicolumn{1}{c|}{15}     & 5  \\
\hline
Segments & \multicolumn{1}{c|}{2015}      & \multicolumn{1}{c|}{821}   & \multicolumn{1}{c|}{1013}   & 109 \\
\bottomrule
\end{tabularx}
\label{tab:train_data}
\end{table}

\begin{table}[h]
\caption{Number of patients and segments per tumour entity in test dataset}
\centering
\setlength\tabcolsep{4pt}
\begin{tabularx}{0.7\columnwidth}{l|ccccc}
\toprule
 & \multicolumn{5}{c}{Test Dataset}\\
\hline
\shortstack[l]{Overall \\ Segments}   & \multicolumn{5}{c}{2657}\\
\hline
Entity       & \multicolumn{1}{c|}{Prostate} & \multicolumn{1}{c|}{Liver} & \multicolumn{1}{c|}{HN} & \multicolumn{1}{c|}{Mamma}  & Lymphnodes\\
\hline
Patients & \multicolumn{1}{c|}{10}       & \multicolumn{1}{c|}{10}    & \multicolumn{1}{c|}{10}     & \multicolumn{1}{c|}{5} & 15  \\
\hline
Segments & \multicolumn{1}{c|}{646}      & \multicolumn{1}{c|}{525}   & \multicolumn{1}{c|}{731}   & \multicolumn{1}{c|}{128} & 627 \\
\bottomrule
\end{tabularx}
\label{tab:test_data}
\end{table}
